\pdfoutput=1

\documentclass[11pt,a4paper]{article}
\usepackage[hyperref]{acl2021}

\usepackage{times}
\usepackage{latexsym}

\usepackage[T1]{fontenc}

\usepackage[utf8]{inputenc}

\usepackage{microtype}

\usepackage{times}
\usepackage{latexsym}
\usepackage{makecell}

\usepackage{soul}
\usepackage{graphicx}
\usepackage{booktabs}
\usepackage{multirow} 
\usepackage{amsmath}
\usepackage{color, xcolor, colortbl}
\definecolor{Gray}{gray}{0.9}
\definecolor{caseblue}{rgb}{0.27451,0.5098,0.70588}
\definecolor{casered}{rgb}{0.27451,0.5098,0.70588}

\newcommand{\tabincell}[2]{\begin{tabular}{@{}#1@{}}#2\end{tabular}}

\usepackage{amssymb}
\usepackage{amsfonts}
\usepackage{tabularx}
\newcommand{\dashrule}[1][black]{%
  \color{#1}\rule[\dimexpr.5ex-.2pt]{4pt}{.4pt}\xleaders\hbox{\rule{4pt}{0pt}\rule[\dimexpr.5ex-.2pt]{4pt}{.4pt}}\hfill\kern0pt%
}
\usepackage{color,xcolor}
\definecolor{LightCyan}{rgb}{0.88,1,1}
\definecolor{babyblue}{rgb}{0.54,0.81,0.94}
\definecolor{airforceblue}{rgb}{0.36, 0.54, 0.66}
\newcommand{\MyColorBox}[2][red]%
{%
    \settowidth{\Width}{#2}%
    \colorbox{#1}%
    {%
        \raisebox{-\DepthReference}%
        {%
                \parbox[b][\HeightReference+\DepthReference][c]{\Width}{\centering#2}%
        }%
    }%
}
\usepackage{todonotes}
\usepackage{url} 
\usepackage{mathtools}
\definecolor{Gray}{gray}{0.9}
\definecolor{LightCyan}{rgb}{0.88,1,1}

\newcommand{\nonl}{\renewcommand{\nl}{\let\nl\oldnl}}
\usepackage{pifont}

\newlength{\Width}
\newlength{\DepthReference}
\newlength{\HeightReference}
\aclfinalcopy 

\newcommand{\aspace}{\hspace{1em}}
\newcommand{\hku}{$^{\heartsuit}$}
\newcommand{\tencent}{$^{\clubsuit}$}
\newcommand{\sdu}{$^{\diamondsuit}$}
\newcommand{\shlab}{$^{\spadesuit}$}

\title{Event Transition Planning for Open-ended Text Generation}
\author{
Qintong Li\hku\thanks{  ~~The majority of this work was done while the first author
was interning at Tencent AI Lab.} \aspace
Piji Li\tencent \aspace
Wei Bi\tencent \aspace
Zhaochun Ren\sdu \aspace 
\textbf{Yuxuan Lai\hku\thanks{  ~~Corresponding author.} \aspace
Lingpeng Kong\hku\shlab \aspace} \\
\hku Department of Computer Science, The University of Hong Kong \\
\tencent Tencent AI Lab 
\sdu Shandong University \\
\shlab Shanghai Artificial Intelligence Laboratory \\
\texttt{qtli@connect.hku.hk} \\
\texttt{\{lipiji.pz, erutan.pkuicst\}@gmail.com} \\  \texttt{victoriabi@tencent.com}, 
\texttt{zhaochun.ren@sdu.edu.cn} \\
\texttt{lpk@cs.hku.hk} \\
}

\begin{document}
\maketitle
\begin{abstract}
Open-ended text generation tasks, such as dialogue generation and story completion, require models to generate a coherent continuation given limited preceding context.
The open-ended nature of these tasks brings new challenges to the neural auto-regressive text generators nowadays. Despite these neural models are good at producing human-like text, it is difficult for them to arrange causalities and relations between given facts and possible ensuing events.
To bridge this gap, we propose a novel two-stage method which explicitly arranges the ensuing events in open-ended text generation. 
Our approach can be understood as a specially-trained coarse-to-fine algorithm, where an event transition planner provides a ``coarse'' plot skeleton and a text generator in the second stage refines the skeleton. 
Experiments on two open-ended text generation tasks demonstrate that our proposed method effectively improves the quality of the generated text, especially in coherence and diversity.
The code is available at: \url{https://github.com/qtli/EventPlanforTextGen}.
\end{abstract}

\section{Introduction}

With the fast development of large-scale pre-trained models, considerable progress has been made in improving the quality of machine generated text \citep{radford2019language,rashkin-etal-2019-towards,zhang2020dialogpt,BrownMRSKDNSSAA20,GuanMFLDH20,BakhtinDGORS21}. Today, machine learning models can do extremely well in generating text that looks human~\citep{clark2021all}. The problem is still far from solved, however, as further reading of the machine-generated text often exposes defects such as self-contradiction and topic drifting \citep{bisk2020experience,gaopllsls20,tan2021progressive,fan2019strategies,dou2021scarecrow,dziri2021neural}. These issues are particularly serious in open-ended text generation tasks (e.g., story completion), where the model is asked to produce a coherent continuation which often involves multiple events, given limited preceding context.

\begin{figure}[!t]
 \centering
 \includegraphics[width=0.48\textwidth]{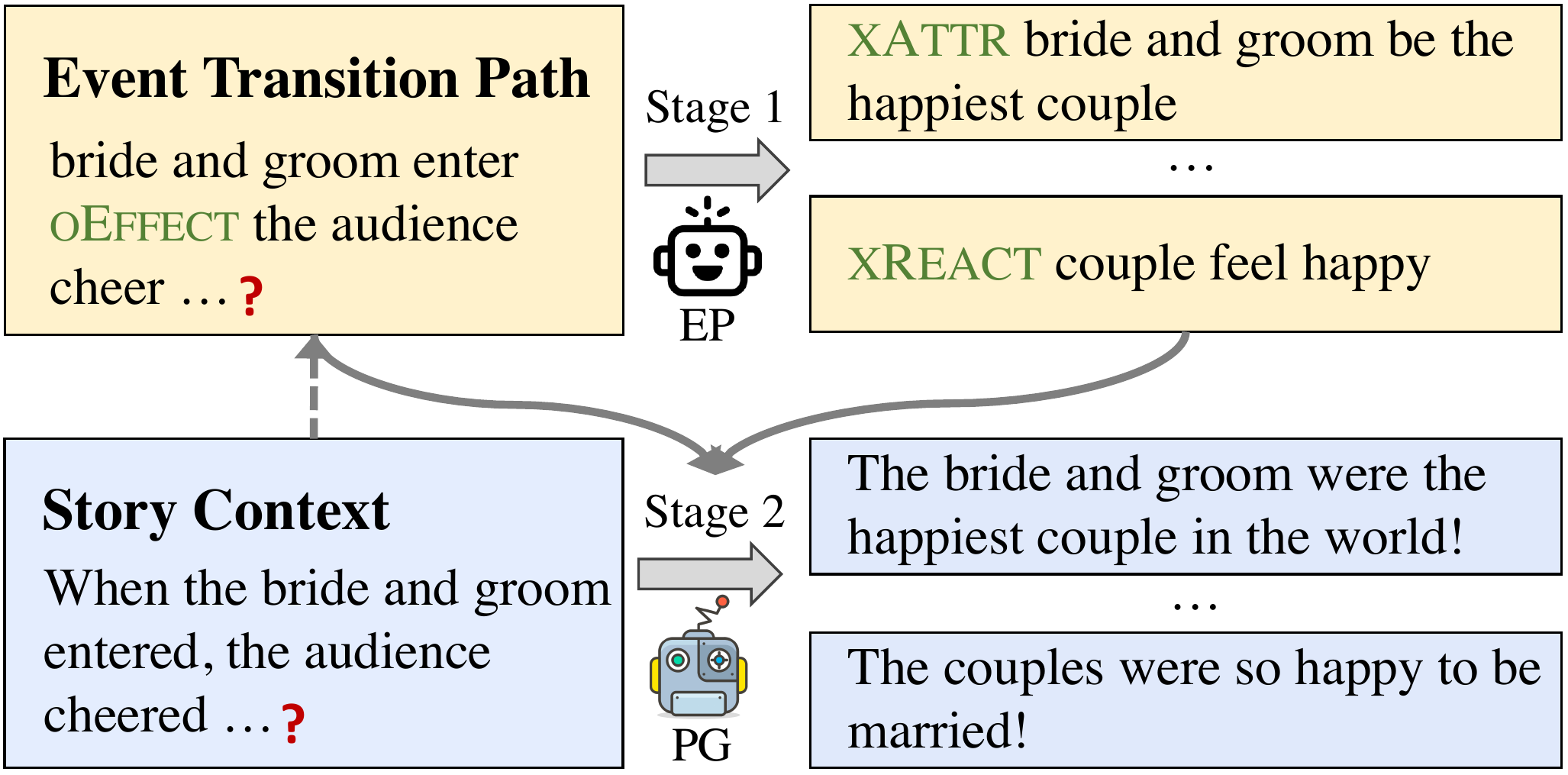}
 \caption{An illustration of our planning based framework in story completion task. Given story context, we extract corresponding \underline{e}vent transition \underline{p}ath, and use model \underline{EP} to develop potential ensuing event transition paths. The planned paths accordingly guide the \underline{p}ath-aware text \underline{g}eneration model \underline{PG}.}
 \label{fig:example}
\end{figure}

\begin{table*}[!ht]
\centering
\setlength{\tabcolsep}{3pt}
\small
\begin{tabular}{p{20mm}|p{55mm}|p{76mm}}
\toprule
& \textbf{Dialogue Generation} & \textbf{Story Completion}\\
\midrule
\textbf{Input Context --- Events} & \tabincell{p{55mm}}{$[1]$ my husband lost a job but i'm hoping he can find a full time job soon. --- \MyColorBox[cyan!10]{\text{my husband lost job}}, \MyColorBox[cyan!10]{\text{I hope he find job}} \\ $[2]$ He will , I have faith. --- \MyColorBox[cyan!10]{\text{I have faith}} \\ $[3]$ thank you so much! --- \MyColorBox[cyan!10]{\text{thank you}}} &  \tabincell{p{78mm}}{$[1]$ John got laid off from his company. --- \MyColorBox[cyan!10]{\text{john get laid off}}\\ $[2]$ He was close to retirement age. ---\MyColorBox[cyan!10]{\text{john is close retirement}}\\ $[3]$ John felt bored and listless his first week of unemployment. --- \MyColorBox[cyan!10]{\text{john feel bored and listless}}\\ $[4]$ John decided to start a business of his own.} --- \MyColorBox[cyan!10]{\text{john decide start business}}\\ 
\midrule
\textbf{Target Output --- Events} & No problem. What kind of work does he do? --- \MyColorBox[cyan!10]{\text{what work he do}} & He now has a flourishing online company. --- \MyColorBox[cyan!10]{\text{john have a}} \MyColorBox[cyan!10]{\text{company}}\\ 
\midrule
\textbf{Event Transition Path} & 
\tabincell{p{55mm}}{\MyColorBox[cyan!10]{\text{my husband lost job}} $\textsc{xAttr}$
\MyColorBox[cyan!10]{\text{i hope he}} \MyColorBox[cyan!10]{\text{find job}}  $\textsc{oReact}$ \MyColorBox[cyan!10]{\text{i have faith}} $\textsc{xReact}$ \MyColorBox[cyan!10]{\text{thank you}} $\textsc{oReact}$ \MyColorBox[cyan!10]{\text{what work he do}}} & 
\tabincell{p{77mm}}{\MyColorBox[cyan!10]{\text{john get laid off}}  $\textsc{xAttr}$ 
\MyColorBox[cyan!10]{\text{john is close to retirement}} 
$\textsc{xReact}$  \MyColorBox[cyan!10]{\text{john feel bored and listless}}  $\textsc{xReact}$  \MyColorBox[cyan!10]{\text{john}} 
\MyColorBox[cyan!10]{\text{decide start business}}  $\textsc{xEffect}$  \MyColorBox[cyan!10]{\text{john have a company}}}  \\
\bottomrule
\end{tabular} 
\caption{Examples of event transition paths acquired from downstream tasks, i.e., dialogue generation
and story completion. Events are marked in \MyColorBox[cyan!10]{\text{blue box}}.} 
\label{tab:task_path}
\end{table*}

To bridge this gap, we propose a two-stage method which explicitly models the event transitions in open-ended text generation. 
Multi-step generation has been adopted to control the generated content at a high level~\cite{LapataD18,ji2020language,xu-2021-global}. Different from previous works that rely on inflexible pattern retrieval, we leverage a generative model as an event transition planner in the first stage to boost the high-level coherence and causalities in open-ended text generation.

Specifically, in stage one, an event transition planner (\S\ref{sec:gep}) outlines a transition path of events starting from the ones extracted from the input context. In stage two, this path is used to ensure a relevant and sound continuation from the actual text generator (\S\ref{sec:ctg}). This method can be understood as a specially-trained coarse-to-fine algorithm, where an event transition planner provides a ``coarse'' plot skeleton and a path-aware text generator refines the skeleton.  Figure \ref{fig:example} shows an illustration of our approach. 

There are two main challenges in this method. First, the planer should produce high-quality and diverse paths that can generalize well to the unseen events at test time. 
For this challenge, we fine-tune a GPT-2 \citep{radford2019language} on a large amount of event paths extracted from commonsense graphs~\citep{sap2019atomic}, as well as from the training set of the specific task, aiming to extrapolate to event sequences that never appeared in these sources with the help of general knowledge stored in the large pre-trained model~\citep{petroni2019language,lee2021neural}. 

For the second challenge, the auto-regressive text generator need to work effectively under the supervision of the even transition path. We thus design an event query layer to absorb information from the planned paths and use the query layer to guide the text generation process.

We validate our method thorough extensive experiments on two standard open-ended text generation tasks, dialogue generation \citep{rashkin2019towards} and story completion \citep{mostafazadeh2016story}. Our two-stage approach outperforms a strong knowledge enhanced GPT-2 baseline \citep{guan-etal-2020-knowledge} in both automatic and human evaluation metrics. Further analysis shows that the improvements of the event transition planning model come in particular from the high-level consistency and diversity in long and difficult generation cases.

\section{Event Transition Path.}
\label{sec:event_t_path}
In this work, the event transition path is defined as an alternating sequence between events and relations, where an \textit{event} is a subject-verb phrase, a \textit{relation} is chosen from a pre-defined label set (e.g., \textsc{oReact} - object reaction; \textsc{xAttr} - subject attribute) of a commonsense atlas~\citep{sap2019atomic}.
Table~\ref{tab:task_path} shows some text examples and their corresponding event transition paths. 
We collect event transition paths from a commonsense atlas \textsc{ATOMIC}~\citep{sap2019atomic}, as well as from the training set of the specific task, to train an event transition planner.

\begin{figure*}[!ht]
    \centering
    \includegraphics[width=0.98\textwidth]{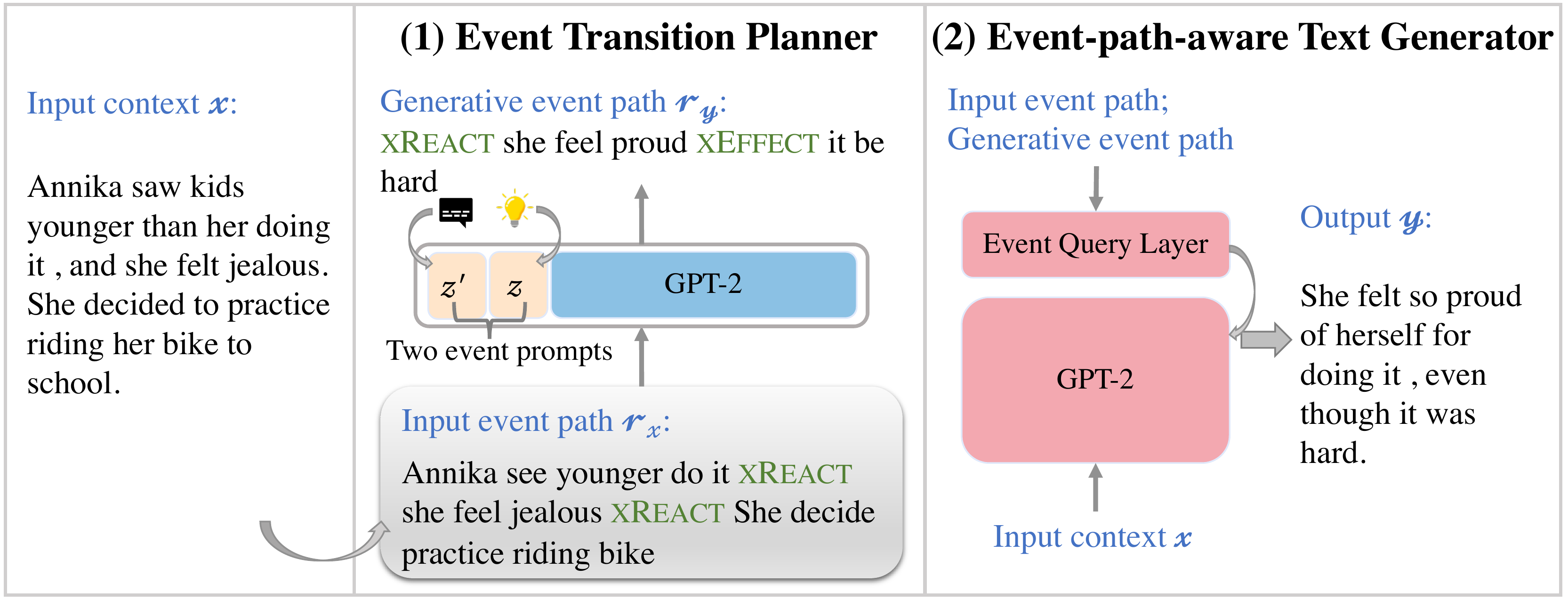}
    \caption{Overall architecture of the proposed coarse-to-fine framework. It consists of two components. (1) \textbf{Event Transition Planner}: given a input context, it first extracts corresponding event path and then generates possible ensuing event path. The planner directly inherits the pre-trained parameters from GPT-2; (2) \textbf{Event-path-aware Text Generator}: another GPT-2-based generator is applied to generate a natural language sentence by attending to input context and explicit event transition path.}
    \label{fig:model}
\end{figure*}

\paragraph{Sampling Paths from ~\textsc{Atomic}.}
We use everyday commonsense atlas \textsc{Atomic}~\citep{sap2019atomic} to acquire plenty event paths. 
\textsc{Atomic} is organized through 9 relations and 877k events (textual descriptions) of inferential commonsense, e.g., if ``PersonX pays PersonY a compliment'', then ``PersonY will likely return the compliment''. It has been demonstrated that \textsc{Atomic} is useful for open-ended text generation tasks, such as story generation~\citep{guan2020knowledge}. 

Besides, to increase the flexibility, 
we introduce a reverse relation (e.g., \textsc{\_xAttr}) for each original relation (e.g., \textsc{xAttr}) so that a sampled path can contain reverse triplets. 
The intuition is that, in open-ended text generation, the narrative maybe in a reverse order.
After explaining the event \texttt{A}, the author may want to introduce the subsequent event \texttt{B}.
Meanwhile, if the author introduce the event \texttt{B} first, she/he may want to describe the event \texttt{A} as an explanation for the reason/motivation.

Finally, we collect sufficient event paths of variant lengths from \textsc{Atomic} via random walking\footnote{The hops of these sampled paths fall in between 1 and 5.}. 
We split the sampled paths into training/validation/test with the ratio of 18:1:1. 
We use these sampled paths to optimize the event transition planner which is responsible for generative event planning (see \S\ref{sec:gep}). 
The statistics of sampled paths are shown in Table~\ref{tab:stats_sampled_path} of Appendix~\ref{app: event_path}. 
We display several examples of the randomly sampled event transition paths in Table~\ref{tab:atomic_path} of Appendix~\ref{app: event_path}.

\paragraph{Extracting Paths from Specific Dataset.}
We use two kinds of event transition paths. A general kind is obtained from random walking on a daily commonsense graph, \textsc{Atomic}, as mentioned above.
Another kind is extracted from the natural language instances of downstream datasets, which is used for the training and prediction stage of task-specific event planning.
For example, given the inputs, ``When the bride and groom entered, the audience cheered'', the extracted event path is ``bride and groom enter \textsc{oEffect} audience cheer''.

In detail, for each sentence, to ensure the extracted events have complete semantics and keep a similar format with the events in \textsc{Atomic}, we use ASER event extractor tool\footnote{\url{https://hkust-knowcomp.github.io/ASER/html/index.html}} to distil events for all sentences of downstream datasets. 
We further predict the relations between these events, linking these isolated events as event transition paths.
Specifically, we train a BERT~\citep{devlin2019bert} classifier using event triples and relations in \textsc{Atomic}.
The sizes of training/validation/test instances are 639,045/35,503/35,502, respectively. We finally achieve a accuracy score of 85\% on the test set for the relation prediction.

\section{Methodology}

We focus on the conditional language modeling problem in open-ended text generation tasks.
Formally, given an input context $\boldsymbol{x}$,
models are required to generate
a sentence $\boldsymbol{y}$
that is consistent with input context and not contradicts itself.

In this work, we propose a two-stage model for the generation process. 
In the first stage, we extract the starting event sequence $\boldsymbol{r}_x$ from the input context and employ the event transition planer to generates subsequent event transition path $\boldsymbol{r}_y$ based on $\boldsymbol{r}_x$. 
In the second stage, the output text is generated from an auto-regressive model conditioning on the path and the preceding context $\boldsymbol{x}$.

Figure~\ref{fig:model} gives an overview of our coarse-to-fine framework for open-ended text generation. In a nutshell, we first fine-tune a GPT-2 on event transition sequences as an event planner (i.e., a conditional generative model for event paths).
This fine-tuning involves event transition sequences extracted from both commonsense graphs and the training set.
We then build a path-aware text generator with an event query layer specifically designed to refer to the planned path when generating the output.

\subsection{Generative Event Transition Planner}
\label{sec:gep}
In this section, we describe the event transition planner which completes the partial event path given certain input context. Pre-trained language models can be good representation learners of relational knowledge~\citep{petroni2019language, bosselut2019comet}. In our model, we choose 
GPT-2~\citep{radford2019language} as the backbone of our event transition planner. 

Specifically, we first fine-tune GPT-2 with large-scaled event transition paths sampled from \textsc{Atomic}~\citep{sap2019atomic}.
After that, we fine-tune the resulting  model in addition on the event transitions extracted from the training corpus, so that the planner is aware of general transitions in the commonsense while focusing on the transitions in the specific domain in the meantime.

In preliminary experiments, we find that directly running a full fine-tuning (i.e., updating all GPT-2 parameters) leads to a drop in the final performance.
We suspect the reason is the full fine-tuning flushes out the original general knowledge from the large-scale pre-training~\citep{ChenWFLW19,lee2019mixout,chen2020recall}.

To overcome this drawback, we prepend a trainable continuous event prompt $\boldsymbol{z}$ to the input path $\boldsymbol{r}=[\boldsymbol{r}_x; \boldsymbol{r}_y]$ of every transformer layer in event transition planner, as prefix-tuning~\citep{LiL20} does. A trainable matrix $\mathbf{U}_\theta$ with parameters $\theta$ is randomly initialized to embed event prompt $\boldsymbol{z}$. The aim is to use parameters $\theta$ introduced by $\boldsymbol{z}$ to store event transition patterns from $\textsc{Atomic}$. Then the representation of each input event transition path $\boldsymbol{r}$ is prompted as $\mathcal{\boldsymbol{r}^\prime}=[\boldsymbol{z}; \boldsymbol{r}]$.
To increase training speed and performance robustness, we apply an additional linear reparameterization function on $\mathbf{U}_\theta$.  
\begin{align}
    \mathbf{U}_\theta = \textit{FFN}_\theta ~(\mathbf{U}^{'}_\theta)\text{,}
\label{eq:repara}
\end{align} 
where $\boldsymbol{U}^{'}_\theta$ is another randomly initialized matrix with smaller dimension, \textit{FFN} is a large feedforward neural network~\cite{vaswani2017attention}. 
We perform gradient updates on the following log-likelihood objective:
\begin{align}
    \max_{\theta}~\log(\boldsymbol{r}_y \mid [\boldsymbol{z}; \boldsymbol{r}_{<y}]) &= \nonumber\\ 
    \sum_{y\in \boldsymbol{z}_{\text{idx}}} \log ~ & \textit{EP}_{\phi,\theta}(\boldsymbol{r}_y \mid \mathbf{h}_{<y}) \text{,} 
\end{align}
where $\phi$ denotes the pre-trained parameters from the backbone LM of event transition planner, $\theta$ denotes the newly introduced parameters for the event prompt, $\boldsymbol{z}_{\text{idx}}$ denotes the index sequence of the event prompt, \textit{EP} is short for \underline{e}vent transition \underline{p}lanner, and $\mathbf{h}_{<y}$ denotes the hidden states calculated by the trainable event prompt matrix and activation layers of the backbone LM:
\begin{align}
    \mathbf{h}_{y} = 
\begin{cases}
\mathbf{U}_{\theta}[y,:],& \text{if } y \in \boldsymbol{z}_{\text{idx}} \text{,}\\
\textbf{\textit{LM}}_\phi(\boldsymbol{r}_y \mid \mathbf{h}_{<y})  & \text{otherwise.}
\end{cases}
\end{align}

Similar to the above event prompting technique, for the paths from downstream dataset, we prepend another event prompt $\boldsymbol{z}^\prime$ to the $\boldsymbol{r}^\prime$ and only optimize the parameters introduced by $\boldsymbol{z}^\prime$. This effectively preserves the newly-learned event transition patterns from \textsc{Atomic} and continuously adapts the event transition planner to different downstream event transition patterns. 

\subsection{Event-path-aware Text Generation}
\label{sec:ctg}
Current state-of-the-art systems for open-ended text generation are based on fine-tuning pre-trained language models with different downstream datasets. Although text generation fluency is usually not a crucial issue nowadays, topic-related mistakes~\citep{dou2021scarecrow} such as off-prompt and self-contradiction are common. 
We therefore integrate the event transition paths produced by the planner into the text generation model via an event query layer using the multi-head attention mechanism ($\textit{MHA}$).

The event query layer is built on top of the stacked transformer layers, aiming to explicitly induce the expected output with event transition paths. The input of the event query layer is the event transition path $\boldsymbol{r}$ given the current input $\boldsymbol{x}$. $\boldsymbol{r}$ not only summarizes the event transition in $\boldsymbol{x}$, also indicates possible event path following $\boldsymbol{x}$. The structure of the event query layer resembles the transformer layer. Its output serves as the key and value vectors in the multi-head attention mechanism, which computes another attention vector $\textit{MHA}(\boldsymbol{r})$. We concatenate two multi-head attention vectors and derive the final event-path-aware attention vector $\mathbf{m}$:
\begin{align}
\mathbf{m} = \textit{MLP}([\textit{MHA}(\boldsymbol{x});\textit{MHA}(\boldsymbol{r})])\text{,}    
\end{align}
where $\textit{MHA}(\boldsymbol{x})$ is the output from the multi-head attention function of the original transformer layer, $\textit{MHA}(\boldsymbol{r})$ is the output from the event query layer. The event-path-aware attention vector $\mathbf{m}$ replaces the original multi-head attention vector $\textit{MHA}(\boldsymbol{x})$ and participates the remaining calculation of the language model.

The optimization of the event-path-aware text generator is the standard cross-entropy objective: $\textit{CrossEntropy}(\boldsymbol{y}_j \mid \boldsymbol{y}_{<j},\boldsymbol{x}, \boldsymbol{r})$.

\subsection{Implementation Details}
We base our event planner and event-plan-aware text generator on pre-trained GPT-2-small models\footnote{\url{https://huggingface.co/gpt2}}.
The event prompt length during training \textsc{Atomic} event transition paths are set to 5 according to pilot study.
We inject and optimize the event query layer on the last layer of the stacked Transformers.
When training the event-path-aware text generator, event path $\boldsymbol{r}_y$ is derived from the ground truth. During inference, $\boldsymbol{r}_y$ is the prediction from event transition planner given the input event transition path $\boldsymbol{r}_x$.
More details are elaborated in Appendix~\ref{app:implement}.

\begin{table*}[!t]
\centering
\small
\setlength{\tabcolsep}{10pt}
\begin{tabular}{llccccc}
    \toprule
    \textbf{Tasks} & \textbf{Methods}    & \textbf{BELU-1}   & \textbf{BLEU-2}   & \textbf{BLEU-4}   & \textbf{DIST-1}   & \textbf{DIST-2}  \\ 
    \midrule
    \multirow{5}{*}{\makecell{Dialogue\\Generation}}
    & \textsc{GPT-2} & 23.43  & 11.50 & ~3.31 & ~~1.57 & ~~4.18 \\
    \cmidrule{2-7} 
    & \textsc{PlanGeneration}~(Ours) & \textbf{26.52}  & \textbf{12.38}   & ~~3.29  & \textbf{~~1.88} & \textbf{~~5.52} \\. 
    & \quad w/o \textsc{Prompt} & 23.58  & 11.85 & \textbf{~~3.58} & ~~1.80  & ~~5.13 \\
    & \quad w/o \textsc{Tuning on Atomic} & 19.82 & ~~7.90 & ~~1.81 & ~~1.16 & ~~2.54 \\
    \cmidrule{2-7} 
    & \textsc{PlanRetrieval} & ~~0.75 & ~~0.14 & ~~0.00 & 13.05 & 39.52 \\
    \midrule
    \multirow{5}{*}{\makecell{Story\\Completion}}
    & \textsc{GPT-2} & 15.98  & ~~7.19 & ~~1.08  & ~~5.53 & 17.44 \\
    \cmidrule{2-7} 
    & \textsc{PlanGeneration}~(Ours) & \textbf{19.51}  & \textbf{~~9.01}   & \textbf{~~1.35}  & \textbf{~~5.83} & \textbf{17.48} \\
    & \quad w/o \textsc{Prompt} & 13.64  & ~~6.14 & ~~1.12 & ~~4.71  & 15.77  \\
    & \quad w/o \textsc{Tuning on Atomic} & 12.74 & ~~4.61 & ~~0.47 & ~~6.08 & 12.27 \\ 
    \cmidrule{2-7} 
    & \textsc{PlanRetrieval} & ~~1.28 & ~~0.15 & ~~0.00 & 11.88 & 37.70 \\
    \bottomrule
\end{tabular}
\caption{Experimental results on event transition planning. For detailed description about the compared models, please refer to \S\ref{sec:event_path_eval}.}
\label{tab:auto_kg}
\end{table*}

\section{Experiments}
We conduct experiments on two open-ended text generation tasks, dialogue generation and story completion, to answer the following questions:

\noindent $\bullet$  \textbf{RQ1}: How to develop a better event transition planner?   

\noindent $\bullet$  \textbf{RQ2}: Whether the integration of event transition paths enhances the open-ended text generation? 

\noindent $\bullet$   \textbf{RQ3}: How do the event transition paths benefit text generation?

\subsection{Evaluated Tasks}

\noindent $\bullet$ \textbf{Story Completion}
requires models to complete a story given the first few sentences.
We evaluated our framework on \textsc{ROCStories}~\citep{mostafazadeh2016story}, which contains 98k five-sentence stories. Our default setting is to predict the last sentence given the first four ones.

\noindent $\bullet$ \textbf{Dialogue Generation}
aims to generate reasonable and human-like responses given the dialogue history. We evaluated our framework on \textsc{EmpatheticDialogues}~\citep{rashkin2019towards} which consists of 25k conversations grounded in pre-specified situations.

\subsection{Event Transition Planning (RQ1)} 
\label{sec:event_path_eval}
We compare our event transition planner, named as \textsc{PlanGeneration}, with fine-tuned pre-trained GPT-2~\citep{radford2019language} and several ablation settings, investigating \textit{how to develop a better event transition planner}. 

Specifically, the compared settings include:

\noindent $\bullet$ \textbf{\textsc{GPT-2}} is a pre-trained GPT-2 model~\citep{radford2019language} directly fine-tuned on the event paths extracted from specific tasks, i.e., dialogue generation or story completion in our work.

\noindent $\bullet$ \textbf{\textsc{PlanGeneration}} is our proposed event planning method, which explores a two-stage fine-tuning on event transition paths from \textsc{Atomic}~\cite{sap2019atomic} and the downstream task, equipping with the proposed event prompting module.

\noindent $\bullet$ \textbf{w/o \textsc{Prompt}} is our proposed method without the event prompting module, but still using the two-stage fine-tuning strategy. 

\noindent $\bullet$ \textbf{w/o \textsc{Tuning on Atomic}} is our proposed method without the first-stage fine-tuning on the event paths extracted from external commonsense atlas \textsc{Atomic}.

\noindent \textbf{$\bullet$ \textsc{PlanRetrieval}} is a retrieval based planning methods, which employs the BM25 ranking function~\citep{robertson1995okapi} to retrieve from the paths extracted from the training sets according to the given context.

\paragraph{Results.} We use \textbf{BLEU}~\citep{papineni2002bleu} and \textbf{DIST}~\citep{li2016diversity} as the automatic metrics to evaluate the generated sentences in terms of the coherence and diversity, respectively. BLEU evaluates $n$-gram overlap between generation and ground truth. BLEU scores will become extremely low for large $n$. We thus experiment with $n \in \{1,2,4\}$. DIST measures the ratio of distinct n-grams to all the generated $n$-grams from the perspective of the generation diversity. For DIST metric, we adopt $n \in \{1,2\}$. The experimental results are shown in Table~\ref{tab:auto_kg}. 
The dataset needed in this section consists of event transition paths sampled from \textsc{Atomic} and extracted from downstream datasets. i.e., \textsc{ROCStories} and \textsc{EmpatheticDialogue}. The details of event transition paths are shown in \S\ref{sec:event_t_path} and Appendix~\ref{app: event_path}.

On both dialogue generation and story completion tasks, our proposed \textsc{PlanGeneration} greatly outperforms baseline GPT-2 on event planning coherence (BLEU) and event path diversity (DIST). 
Specifically, on two downsteam tasks, our event transition planner \textsc{PlanGeneration} surpasses the fine-tuned GPT-2 by 3.09 and 3.53 on BLEU-1, 0.31 and 0.30 on DIST-1. This improvement indicates that (1) the two-stage event prompting module could endow event transition planner powerful abilities on predicting the ensuing event paths; (2) enhanced with the large-scale event transition patterns from \textsc{Atomic}, our event transition planner becomes more creative and produces more diverse outcomes.

Considering the ablation settings, without tuning on \textsc{Atomic}~(w/o \textsc{Tuning on Atomic}) or without the event prompting module~(w/o \textsc{Prompt}), the method performs worse on both tasks and across almost all metrics. 
The limited performance of w/o \textsc{Tuning on Atomic} suggests the necessity and effectiveness of learning general event transition patterns from \textsc{Atomic} before optimizing on task-specific event paths. 
Tuning on \textsc{Atomic} event patterns could make event transition planner get familiar with the event-path-like language and generalize well on unseen event patterns.
Compared to ablation model w/o \textsc{Prompt}, \textsc{PlanGeneration} is comparatively more effective. This is because when optimizing on event paths of target tasks, the proposed event prompt protects the parameters of pre-trained language model from drastic change when training with event transition paths. 
This comparison confirms our intuition that event prompting module could improve event planning performance without destroying the eventual commonsense stored in pre-trained parameters. It provides a more flexible approach to blend the event transition patterns in both \textsc{Atomic} and specific tasks with the pre-trained GPT-2 model.

We also attempt a variation of our \textsc{PlanGeneration} method, i.e., \textsc{PlanRetrieval}.
We can see that the BELU scores of \textsc{PlanRetrieval} are substantially lower that the generation based methods. The main reason is that the target event paths are flexible, infinite, and task-related. Many transition patterns are not seen in the training data or external commonsense graph.

\begin{table*}[t!]
\centering
\small
\setlength{\tabcolsep}{8pt}
\begin{tabular}{llccccc}
    \toprule
    \textbf{Tasks} & \textbf{Models} & \textbf{BLEU-1} & \textbf{BLEU-2}  & \textbf{BLEU-4} &  \textbf{DIST-1} & \textbf{DIST-2}    \\ 
    \midrule
    \multirow{4}{*}{\makecell{Dialogue\\Generation}}
    & \textsc{GPT-2} & 16.07 & 6.41 & 2.13 & 2.06 & 7.70 \\
    & {\textsc{GPT-2-CS-FT}}
    (\citeauthor{guan2020knowledge}) & 16.43 & 6.83 & 2.31 & 2.16 & 8.28  \\
    & \textsc{R-EP-PG} & 16.68 & 6.71 & 2.27 & \textbf{2.21} & \textbf{8.44} \\
    & \textsc{EP-PG}~(Ours) & \textbf{16.74} & \textbf{6.94} & \textbf{2.39} & 2.19 & 8.25 \\
    \midrule
    \multirow{4}{*}{\makecell{Story\\Completion}}
    & GPT-2 & 25.03 & 9.58 & 2.70 & 8.38 & 31.33  \\
    & {\textsc{GPT-2-CS-FT}}
    (\citeauthor{guan2020knowledge}) & 25.09 &  9.64 & 2.72 & 8.07 & 30.68 \\
    & \textsc{R-EP-PG} & 24.72 & 9.27 & 2.63 & 7.01 & 26.49 \\
    & \textsc{EP-PG}~(Ours) & \textbf{25.47} & \textbf{9.71} & \textbf{2.74} & \textbf{8.99} & \textbf{34.48} \\
    \bottomrule
\end{tabular}
\caption{Results of experiments on open-ended text generations. For detailed information about each compared model, please refer to \S\ref{sec:exp_eg}.}
\label{tab:auto_eval_kgen}
\end{table*}

\subsection{Event-path-aware Text Generation (RQ2)}
\label{sec:exp_eg}

In this section, we compare our overall framework \textsc{EP-PG} with several baselines to investigate \textit{whether the integration of generative event transition paths benefits the open-ended text generation}.

We consider the following settings:

\noindent $\bullet$ \textbf{\textsc{GPT-2}} is a pre-trained GPT-2 model~\citep{radford2019language} fine-tuned on the task-specific dataset.

\noindent $\bullet$ \textbf{\textsc{GPT-2-CS-FT}} is a commonsense-enhanced GPT-2 model. By following \citet{guan2020knowledge}, we conduct a first-stage post-training on the \textsc{Atomic} commonsense triples and then fine-tuning on task-specific dataset.

\noindent $\bullet$ \textbf{\textsc{EP-PG}} is our proposed framework, which is a fine-tuned GPT-2 model integrated with the event transition path produced from event transition planner \textsc{PlanGeneration} via an event query layer. 

\noindent $\bullet$ \textbf{\textsc{R-EP-PG}} is another version of \textbf{\textsc{EP-PG}} to explore the proposed event query layer. The input event transition paths are produced by \textsc{PlanRetrieval} in a retrieval way.

\paragraph{Results.}
We consider the same evaluation metrics as in \S\ref{sec:event_path_eval}. As demonstrated in Table~\ref{tab:auto_eval_kgen}, \textsc{EP-PG} achieves the most satisfying performance among all settings on both tasks\footnote{P-value $<$ 0.05 on BLEU-1, according to \citet{sigf}.}. Integrated with the explicit guidance of the event transition paths, \textsc{EP-PG} produces more accurate open-ended generations with higher diversity.  

Particularly, our proposed framework \textsc{EP-PG} consistently and significantly improves GPT-2 baseline for all tasks on content quality (BLEU) and diversity (DIST), showcasing the advantage of injecting event query layer on fine-tuned GPT-2.
Without the explicit modeling of event transition paths, \textsc{GPT-2-CS-FT} which post-trains on commonsense triples only obtains a slight improvement or even performs comparable with GPT-2 model. 
\textsc{EP-PG} further improves generation performance from \textsc{GPT-2-CS-FT} across all metrics on the two tasks, highlighting the efficacy of long-range event planning via an additional event query layer.

Particularly remarkable are the relative differences between \textsc{R-EP-PG} and \textsc{EP-PG}. Although \textsc{R-EP-PG} manages to bring generations more diversity, but in most cases, \textsc{EP-PG} is more effective on content planning and informativeness due to generative event transition patterns in higher qualities.  
Moreover, \textsc{R-EP-PG} performs even worse than GPT-2 on story completion. This implies that low-quality event paths even damage the generations. Thus, a reliable event path is a key guarantee for effective downstream text generation.

\begin{table*}[t]
\centering
\small
\setlength{\tabcolsep}{7.5pt}
\begin{tabular}{llcrcccrcc}    
\toprule
    \multicolumn{1}{c}{\multirow{2}{*}{\textbf{Tasks}}} &
    \multicolumn{1}{c}{\multirow{2}{*}{\textbf{Models}}} & \multicolumn{4}{c}{~~~~~~\textbf{Coherence}} & \multicolumn{4}{c}{~~~~~~\textbf{Diversity}} \\
    & &~~~~~~\textbf{Win} & \textbf{Lose} & \textbf{Tie} & $\kappa$ & ~~~~~~\textbf{Win} & \textbf{Lose} & \textbf{Tie} & $\kappa$\\
    \midrule
    \multirow{3}{*}{\makecell{Dialogue\\Generation}} &
    Ours vs. GPT-2 &~~~~~~45\% & 11\% & 44\% & 0.290 &~~~~~~  71\% & 10\% & 19\% & 0.226  \\
    & Ours vs. GPT-2-CS-FT &~~~~~~34\% & 10\% & 56\% & 0.286 &~~~~~~ 54\% & 7\% & 39\% & 0.288  \\
    & Ours vs. \textsc{R-EP-PG}
    &~~~~~~32\% & 8\% & 60\% & 0.472 &~~~~~~ 67\% & 11\% & 22\% & 0.291 \\
    \midrule
    \multirow{3}{*}{\makecell{Story\\Completion}} & 
    Ours vs. GPT-2 &~~~~~~45\% & 12\% & 42\% & 0.397 &~~~~~~ 59\% & 10\% & 31\% & 0.220  \\
    & Ours vs. GPT-2-CS-FT &~~~~~~47\% & 17\% & 36\% & 0.387 &~~~~~~ 56\% & 17\% & 27\% & 0.210  \\
    & Ours vs. 
    \textsc{R-EP-PG}
    &~~~~~~43\% & 17\% & 40\% & 0.393 &~~~~~~ 61\% & 6\% &  33\% & 0.340 \\
    \bottomrule
\end{tabular}
\caption{Manual evaluation results on downstream text generation.  The scores indicate the percentages of Win, Lose or Tie when our model is compared with other baselines. $\kappa$ denotes Fleiss' kappa (all are \textit{fair agreement} or \textit{moderate agreement}).}
\label{tab:huam_eva}
\end{table*}

\subsection{Analysis: Event Transition Planning for Different Generation Scenarios (RQ3)}
To further investigate \textit{how do the event paths benefit text generation}, we analyse the effectiveness of event paths on differently difficult levels of generation, i.e., token-level and sentence-level. 

\paragraph{Token-level.} We first separate the test set into 5 groups according to the averaged target sentence lengths, and then observe the improvements of our proposed \textsc{EP-PG} over GPT-2 on BLUE-1 score. We find that our framework gains more on the longer instances in both story completion (from 0.4 to 1.3 on instances with more than 15 non-stop-words) and dialogue generation (from 0.3 to 0.9 on instances with more than 5 non-stop-words). We argue that the longer targets imply more sophisticated upcoming event transitions, where the guidance from the event transition planner becomes more important.

\begin{figure}[!t]
\centering
 \includegraphics[width=0.4\textwidth]{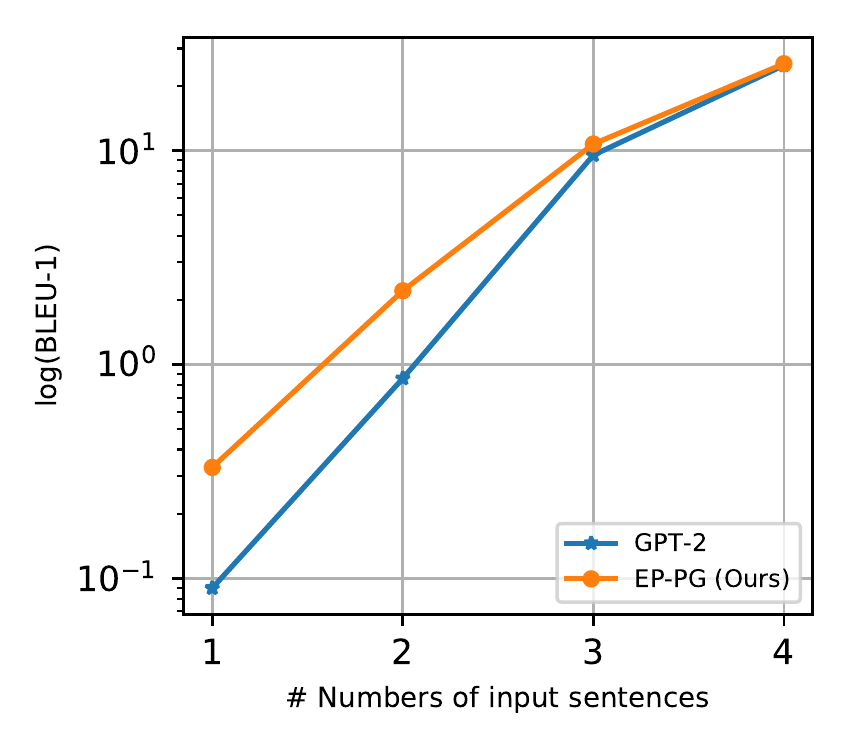}
 \caption{The log of BLEU-1 scores on story completion with different numbers of sentences as input.} 
 \label{fig:grow_rate_of_bleu} 
\end{figure}

\paragraph{Sentence-level.} For story completion on five-sentence story dataset \textsc{ROCStories}, we further conduct experiments on \textsc{EP-PG} with various input sentences and output sentences, i.e.,  the numbers of input (output) sentence are 1 (4), 2 (3), 3 (2), and 4 (1), respectively. Figure~\ref{fig:grow_rate_of_bleu} shows that, compared to GPT-2, the relative improvement proportion of \textsc{EP-PG} is nearly doubled on the most difficult setting where there is only one sentence as input. This improvement is much larger than the easiest situation where 4 sentences are input to the model.
Despite less input context, \textsc{EP-PG} with event transition planning manages to performs better with smaller performance drop.

\subsection{Human Evaluation}
We set up a human evaluation as a complementary evaluation beyond automatic evaluation. For both tasks, we randomly select 100 samples from test set. For each sample, we compare three pairs of models: \textsc{EP-PG} versus \textsc{GPT-2}, \textsc{GPT-2-CS-FT}, and \textsc{R-EP-PG}. Each comparison is rated by three crowd workers, who are asked to give a preference (win, lose or tie) from two perspectives:

\noindent $\bullet$ \textbf{Coherence.} It indicates whether the inference is natural, relevant, and follows logically from the given context.

\noindent $\bullet$ \textbf{Diversity.} Particularly, for baseline models, we use beam search decoding with a top-$k$ ($k=5$) sampling scheme~\citep{fan2018hierarchical} and a softmax temperature $\tau$ ($\tau=0.7$) to generate three inferences per sample. For our method \textsc{EP-PG}, its event transition planner first predicts three paths via the same beam decoding, then its text generator uses greedy decoding based on the generated three paths to produce three inferences per sample. 
During pair-wise comparison, we ask annotators to evaluate which model's predictions contain more reasonable and coherent event transition patterns.

The two aspects are independently evaluated and results are shown in Table~\ref{tab:huam_eva}.
According to human evaluations, our proposed \textsc{EP-PG} significantly outperforms compared baselines in terms of both criteria on the test set of all datasets.
Overall inter-rater agreement measured by Fleiss’ kappa~\citep{fleiss1971measuring} and all the results show fair agreement~($0.2 \leq k \leq 0.4$) or moderate agreement~($0.4 \leq k \leq 0.6$).
The results indicate that explicit incorporating event transition patterns yields significant improvement in generating coherence texts given the input context.
Specifically, with guidance from different event transition paths, our method could produce more diverse and reasonable inferences.

\subsection{Qualitative Study}

Table~\ref{tab:case_study} illustrates how our model tends to produce more contentful and coherent predictions compared to the other systems. In this story completion case, the generated event path successfully captures the correlations between \textit{working out} and \textit{pass physical test}, which further helps our model produce the most reasonable output, \textit{Alex was able to pass the physical exam}. 
For the baseline without commonsense knowledge, GPT-2, is instead not related to the core context \textit{failed the physical assessment}. Tuning on commonsense atlas \textsc{Atomic}, GPT-2-CS-FT produces informative inference but contradicts the context.
The retrieval-based model \textsc{R-EP-PG} searches a related event transition \textit{police officer}. However, its flexibility is limited by search space and cannot maintain a long-range event path, which is easy to produce hallucinated inference. More case analysis
are stated in the Appendix~\ref{app:case}.

\begin{table}[!t]
    \small
    \centering
    \setlength{\tabcolsep}{0.7mm}{
    \begin{tabular}{p{7.5cm}}
        \toprule
        \textbf{Story Context}: \\
          \quad Alex was in training to be a police officer.\\ 
          \quad  He was not in the best shape. \\ 
          \quad Alex failed the physical assessment.  \\ 
          \quad Alex started working out. \\
        \midrule
        \textbf{Golden Event Path: }
        \\\quad \textsc{xEffect} he take the test again \textsc{xEffect} he pass \\
        \textbf{Retrieved Event Path: }
        \\\quad wants to be best police officer \textsc{xWant} tells person to stop \\
        \textbf{Generated Event Path: }
        \\\quad \textsc{xEffect} Alex able get good shape \textsc{xEffect} Alex able pass physical test \\
        \midrule
        \textbf{Reference:} \\\quad He took the test again and passed . \\
        \textbf{GPT-2:}  \\\quad Alex was able to get a good job. \\
        \textbf{GPT-2-CS-FT:} \\\quad Alex made the squad. \\
        \textbf{\textsc{R-EP-PG}:} \\\quad Alex was able to become a police officer. \\
        \textbf{\textsc{EP-PG}:} \\\quad Alex was able to pass the physical exam. \\
        \bottomrule
    \end{tabular}}
    \caption{Case study on story completion. The three sections from top to bottom are the input context, the event transition plans, and inferences from our model and baseline models, respectively.}
    \label{tab:case_study}
\end{table}

\section{Related Work}

Recent advances in pre-trained language models have resulted in impressive performances on open-domain text generation, such as story
completion~\citep{see2019massively,YaoPWK0Y19,fan2019strategies,ippolito2020toward}, dialogue generation~\citep{rashkin2019towards,zhang2020dialogpt,li2020empirical,vulic2021convfit}, question generation~\citep{ChengLLZLLZ20,wang2021multi}, and so on. 
For example, 
in dialogue generation, \citet{zhang2020dialogpt} design a trainable generative pre-trained transformer by training an autoregressive language model on large-scale Reddit context-response pairs 
with a maximum mutual information scoring function to improve diversity. 
\citet{goldfarb2020content} integrate semantic role labels and prompts into pre-trained BART~\cite{lewis2020bart} during fine-tuning for prompt based story telling.
In this paper, we focus on story completion and dialogue generation and build a generative coarse-to-fine method to generate open-ended text with explicit event transition paths.

Despite the success of generative pre-trained language models on a series of open-ended text generation tasks, they still suffer in maintaining coherence throughout multiple sentences due to the left-to-right word-by-word generation style~\citep{fan2019strategies,yu_etal2020}. 
To alleviate this problem, one research direction adopts coarse-to-fine progressive text generation~\cite{tan2021progressive}. This generation paradigm has been studied in many text generation systems for specific tasks, such as data-to-text generation~\cite{MoryossefGD19,PuduppullyL21}, storytelling~\cite{goldfarb2020content,orbach2020facts2story}, and dialogue generation~\cite{XuLWN0C20}. Our work adopts a generative event transition planner that is trained on a large amount of event transition paths, aiming to arrange the ensuing events in open-ended text generation.

Another research direction incorporates external entities to guide the open-ended text generation~\citep{guan2019story,ZhangLXL20,dziri2021neural,li2020knowledge,penglwe21}. \citet{ji2020language} and \citet{xu2020controllable} retrieve entities from knowledge bases to control the generated content.
However, the retrieval-based methods also suffer from the sparsity problem and the domain shift between external sources and downstream tasks~\citep{wang2020connecting}.
\citet{guan2020knowledge} integrate entity relations into pre-trained language model via additional tuning on entity triples. 
Even with such specialized learning, the resulted model still often stuck in logical errors or repeats pieces of narratives~\citep{guan2020knowledge,penglwe21}. 
This phenomenon demonstrates the need for an intact inductive bias on organizing event transition patterns for open-ended text generation.
Different from using event triples as additional training instances, our method explicitly maintains generative event transition paths to make the generation process more explainable and improve the coherence.

\section{Conclusion}
In this paper, we propose a novel two-stage method to improve high-level consistency and diversity in open-ended text generation.
We design a special-trained event transition planner to explicitly arrange the ensuing events and introduce an event-path-aware text generator to exploit the event transition guidance for language generation. 
We investigate two open-ended text generation tasks, i.e., story completion and dialogue generation.
Thorough experiments demonstrate that the explicit arrangement of event transition path indeed facilitate models to generate more coherent and diverse text in open-ended scenery.
Besides, with the proposed event prompt and event query layer, our method could be extended to any other language models and open-ended generation tasks. 
A future line of investigation is to explore the effect of the proposed method on other open-ended tasks, such as commonsense question answering.

\section*{Acknowledgments}
We thank the anonymous reviewers whose suggestions helped clarify this work. This research is supported in part by the National Natural Science Foundation of China (Grant No. 62106105, 61902219), the Shanghai Committee of Science and Technology, China (Grant No. 21DZ1100100), and the Tencent AI Lab Rhino-Bird Focused Research Program.

\bibliographystyle{acl_natbib}
\bibliography{anthology,custom}

\appendix

\section{Event Transition Path}
\label{app: event_path}
The statistics about event transition paths sampled from~\textsc{ATOMIC} are shown in Table~\ref{tab:stats_sampled_path}. We display several examples of the sampled event transition paths in Table~\ref{tab:atomic_path}.

\section{Implementation Details}
\label{app:implement}
For all the systems, including the event transition planner and text generator in our proposed method, we employ the small version of GPT-2 model\footnote{\url{https://huggingface.co/gpt2}} which is a Transformer with 12-head, 12-layer, and hidden size of 768.
The total parameter scalse is 117M.
We use pre-trained GPT-2 Byte Pair Encoding (BPE) tokenizer with an extended vocabulary of 50,282 tokens to tokenize texts.

The event prompt length during training \textsc{Atomic} event transition paths, \textsc{EmpatheticDialogues} paths, and \textsc{ROCStories} paths are 5, 5, and 10, respectively.
The dimension of the randomly initialized smaller matrix $\mathbf{U}^\prime_\theta$ in Eq.\ref{eq:repara} is 512.

The batch size is 128 using AdamW optimizer~\citep{loshchilov2018decoupled} with a learning rate of 5e-5. 
We select the best checkpoint according to the perplexity on the development set of each task and apply early stopping on training where the patient value is set to 2.
We adopt the pre-trained BERT-base model\footnote{\url{https://huggingface.co/bert-base-uncased}} to train the event relation classifier.
All experiments are implemented by PyTorch framework~\citep{paszke2017automatic} and run on NVIDIA V100 GPUs. 
The training time of the event transition planner and event-path-aware text generator are less than 5 hours and 3 hours with 8 GPUs.

\begin{table}[!t]
\centering
\small
\setlength{\tabcolsep}{8pt}
\begin{tabular}{c|c|c|c}
  \toprule
  \textbf{Total} &
  \textbf{Training} &
  \textbf{Validation} &
  \textbf{Test} 
  \\
  \midrule
  4,016,468 & 3,614,981  & 200,752 & 200,735 \\
  \bottomrule
\end{tabular}
\caption{
Numbers of the sampled event transition paths from \textsc{Atomic}.
}
\label{tab:stats_sampled_path}
\end{table}

\begin{table}[!t]
\centering
\setlength{\tabcolsep}{3pt}
\small
\begin{tabular}{p{75mm}}
\toprule
\multicolumn{1}{c}{\textbf{Sampled Event Transition Paths of Variant Lengths}} \\ \midrule
$[1]$ \quad PersonX earns a bachelor's degree \quad \textsc{xWant} \quad PersonX wants to find a good job \\
$[2]$ \quad PersonX asks PersonY to join \quad \textsc{oWant} 	\quad PersonY wants to be friends \quad \textsc{xReact} \quad PersonY feels loved  \\
$[3]$ \quad PersonX is inebriated \quad \textsc{\_xAttr} \quad  PersonX loses control of PersonX's car \quad \textsc{xReact} \quad  PersonX feels scared  \quad \textsc{\_oReact} \quad PersonY takes PersonX by force \quad \textsc{xReact} \quad PersonY feels triumphant \\
\bottomrule
\end{tabular}
\caption{Event transition paths sampled from daily commonsense reasoning atlas \textsc{Atomic}~\citep{sap2019atomic}.}
\label{tab:atomic_path}
\end{table}

\section{Case Study}
\label{app:case}
We qualitatively analyze our model predictions and find that although the proposed model outperforms the state-of-the-art baselines, many of predictions are still wrong. Table~\ref{tab:good_bad_cases} shows several satisfying and unsatisfying predictions on the two datasets. 
One significant error originates from the weak alignment between event transition path and final prediction. 
For example, in the second case, despite ``\textsc{xEffect} tommy be happy'' is imperfect, the prediction ``bought it'' do not convey its information and makes co-reference mistake~(the expected output is ``bought them''). Another serious error type is event transition hallucination, where both the predicted event path and its corresponding inference fail to maintain the logic coherence, such as the fourth case. 
These problems could be alleviated if we design a better format of event transition path which is easier to learn or improve the relation modeling between events and sentences.

\begin{table*}[h]
\centering
\setlength{\tabcolsep}{3pt} 
\small
\begin{tabular}{c|p{6.5cm}|p{6.5cm}}
    \toprule
    & \textbf{Input} & \textbf{Corresponding Event Transition Path}  \\ 
    \midrule
    \multirow{9}{*}{\makecell{\textbf{Good}\\\textbf{Case}\\\textbf{on}\\\textbf{Story}\\\textbf{Completion}}}
    & \textbf{Context}:  & \\
    & Our granddaughter is two. & Our granddaughter be two \\
    & Today she went to the doctor for a blood draw. & \textsc{xEffect} she go doctor for draw  \\
    & She did very well.  & \textsc{xEffect} she do well  \\
    & Our daughter sent a photo of her licking a lollypop afterward. & \textsc{oReact} we daughter send a photo \textsc{xEffect} she lick lollypop    \\ 
    & \textbf{Target}: &  \\
    & We were very proud of her. & \textsc{oEffect} we be proud\\
    & \textbf{Prediction}: & \\
    & We were amused by the photo. & \textsc{xEffect} we get good photo \textsc{xEffect} we be happy \\
    \midrule
    \multirow{9}{*}{\makecell{\textit{\textbf{Bad}}\\\textbf{Case}\\\textbf{on}\\\textbf{Story}\\\textbf{Completion}}}
    & \textbf{Context}:  & \\
    & Tommy wanted to buy a new computer. & tommy want buy new computer  \\
    & After some research he decided to build a PC himself. &  \textsc{xEffect} After research decide build PC  \\
    & He found a site that spelled out compatible parts lists. &  \textsc{xEffect} he find site   \\
    &  He shopped around for the cheapest parts he could get.  & \textsc{xEffect} he shop around part \textsc{\_xEffect} he could get \\
    & \textbf{Target}: &  \\
    & The PC he made was more powerful than computers twice its price. & \textsc{xEffect} he make powerful computer \\
    & \textbf{Prediction}: &  \\
    & He finally found the perfect parts list and \textcolor{airforceblue}{bought it}. & \textsc{xReact} he find part \textcolor{airforceblue}{\textsc{xEffect} tommy be happy}  \\
    \midrule
    \multirow{7}{*}{\makecell{\textbf{Good}\\\textbf{Case}\\\textbf{on}\\\textbf{Dialogue}\\\textbf{Generation}}}
    & \textbf{Context}:  & \\
    &Hi, I joined a firm 6 months ago and then I got a promotion for Junior Manager. & i join firm ago  \textsc{xEffect} i get promotion for manager  \\
    & \textbf{Target}: & \\
    &Congratulations. That sounds like the fast track.  & \textsc{oReact} that like fast track\\
    & \textbf{Prediction}: &  \\
    &That's awesome! I bet you are excited!  & \textsc{oReact} that be great \textsc{xEffect} you must be proud\\
    \midrule
    \multirow{5}{*}{\makecell{\textit{\textbf{Bad}}\\\textbf{Case}\\\textbf{on}\\\textbf{Dialogue}\\\textbf{Generation}}}
    & \textbf{Context}:  & \\
    & I got my four year old daughter her first tricycle yesterday.  &  i get my daughter tricycle yesterday\\
    & \textbf{Target}: &  \\
    & thats so sweet of you.  & \textsc{oReact} that sweet\\
    & \textbf{Prediction}: & \\
    &Wow, that's a lot of fun. \textcolor{airforceblue}{What kind of tricycle?} & \textsc{oEffect} that be great \textcolor{airforceblue}{\textsc{xEffect} what be tricycle} \\
    \bottomrule
\end{tabular}
\label{tab:multicol}
\caption{Summary table of issues found through a qualitative analysis of our model predictions. The errors that occur in our model predictions are colored in \textcolor{airforceblue}{blue}.}
\label{tab:good_bad_cases}
\end{table*}

\end{document}